# Rule Based Stemmer in Urdu

Vaishali Gupta[#1], Nisheeth Joshi[#2], Iti Mathur[#3]

[#]*Apaji Institute, Banasthali University, Rajasthan, India*

[1]`vaishali.gupta77@gmail.com`
[2]`nisheeth.joshi@rediffmail.com`
[3]`mathur_iti@rediffmail.com`

*Abstract*— **Urdu is a combination of several languages like Arabic, Hindi, English, Turkish, Sanskrit etc. It has a complex and rich morphology. This is the reason why not much work has been done in Urdu language processing. Stemming is used to convert a word into its respective root form. In stemming, we separate the suffix and prefix from the word. It is useful in search engines, natural language processing and word processing, spell checkers, word parsing, word frequency and count studies. This paper presents a rule based stemmer for Urdu. The stemmer that we have discussed here is used in information retrieval. We have also evaluated our results by verifying it with a human expert.**

## I. Introduction

Stemming is a process in which affixes are separated from its root word. To understand this phenomenon, let us consider a few words: Healthy, Healthier and unhealthy, it contains some affixes like 'un', 'y' and 'ier'. The common root or stem word is 'Health'. This common root is morphologically related to various variant words. Similarly in Urdu, we have بےیمان (*bey-imaan*), ایماندار(*imaan-daar*), ایمانداری(*imaan-daari*). Here affixes are بے (*be*), دار (*daar*) and داری (*daari*) and its common root word is ایمان (*imaan*).

Stemmer is an algorithm through which we can reduce a word to its stem. It is an important phenomenon for development of natural language processing applications. It can be used in spell checking, machine translation and information retrieval systems. In this paper, we present a Rule Based Stemmer for Urdu. There are many challenges in Urdu language to deal with. One of the foremost reason is that Urdu is a weakly inflectional language and due to its complex and rich morphology and its diverse nature, stemming in Urdu is quite challenging. For Example, a word مشہور(*mashhoor*) (Singular) and مشاہیر (*mashaheer*)(Plural). These words don't present any inflection due to its complex morphology. Here root word cannot be separated from the word. These type of words are considered as exceptions and are termed as part of irregular morphology. Examples of regular morphology are بدمزاج (*bad-mizaj*), بدمزاجی (*bad-mizaj-i*) and مزاج آشنا(*mizaj-aashna*). Here we can extract the common root 'مزاج' (*mizaj*) from these words.

Stemming process has some errors associated with it. These errors are Under Stemming and Over Stemming errors. In Under Stemming, a word is misinterpreted and its affixes are not removed. For example, from word پیشگی (*peshgi*) suffix extracted ی (*i*) and root word is (*peshag*). Here پیشگ (*peshag*) is not a valid stem or root word. In place of پیشگ (*peshag*), root word should be given as پیش (*pesh*) and suffix گی (*gi*) is extracted from word. In Over Stemming, affixes which were not supposed to be removed are removed. For example, from word بدمعاش (*badmaash*) prefix extracted بد (*bad*) and root word is ماش (*maash*). Here, ماش (*maash*) is not a valid root word. For this word we do not need to extract prefix and can display the word as it is بدمعاش (*badmaash*). In case of over stemming, extra affixes are removed.

## II. Related Work

In the field of natural language processing, first stemmer was developed by Lovins [1] in 1968. She proposed 260 rules for stemimng out words of English language. Porter [2] developed a stemmer on the basis of separating affixes. This stemmer performed stemming process in five steps. In the first step, inflectional suffixes were handled, through the next three steps derivational suffixes were handled and in the last step recoding was done.

Khoja and Garside [3] developed an Arabic Stemmer which they called "superior root based stemmer". This stemming algorithm separated prefix, suffix and infixes and then matched the word with the affixes and pulled out the word. This algorithm faced several problems especially with nouns. Al-shammari and Lin [4] presented an Educated Text Stemmer (ETS). It was a very simple, dictionary free and efficient Stemmer which decreased the stemming errors and required lesser storage and lesser processing time. Akram et al [5] proposed a Stemmer for Urdu language. Their work was restricted to Urdu language which meant that root word was not extracted from other languages like English, Persian and Arabic. This stemmer separated the prefix and suffix from word. This system was unable to handle words having infixes. Khan et al [6] presented several challenges for developing a rule based Urdu Stemmer. They discussed some rule based stemming algorithms for English, Arabic, Persian and Urdu language. By their proposed rule based stemming algorithm, they separated suffixes and prefixes and even proposed some pattern for handling them.

Ameta et al [7] proposed a Guajarati stemmer which was an inflectional stemmer which used 167 rules to stem out the words. Through their experimwnts, they reported an accuracy of 91.7% and used this system in factored translation model for Gujarati-Hindi MT System [8]. Pal et al [9] proposed a Hindi Lemmatizer which used 112 rules to extract the root word from an inflected word. They reported an accuracy of 91% for this system. Mishra and Prakash [10] proposed an effective stemmer for Hindi language. This Stemmer resolved the problem of over stemming and under stemming. This proposed stemmer was totally based on Devnagri script and it gave the accuracy of 91.59%. Husain [11] proposed an unsupervised approach to develop a stemmer for Urdu and Marathi language. In this approach, he proposed two ways for suffix generation: frequency based suffix stripping algorithm and length based suffix stripping algorithm. Pal et al [12] proposed a rules based lemmatizer which has 124 rules for generation of root word. This system could handle both suffixes as well as prefixes and showed an accuracy of 89%.

III. Morphology of Urdu

Urdu is quite similar to other Indo-Aryan languages. It has a strong Perso Arabic script, which is written from right to left. Urdu is a free word order and weakly inflectional language. It has a complex and rich morphology. In Morphology, words can be divided into lexical categories with the help of inflection. Nouns, verbs, adjectives and adverbs are example of some common lexical categories. In Urdu morphology nouns, verbs, adjectives and adverbs can also have multiple inflectional forms. It also has a gender, number, person, tense etc based infected forms. As like Hindi, Urdu can have multiple postpositions added with noun or verb or adjective. For Example: خشمزاج(*khushi*), خوشحالی(*khushhaali*), (*khushmizaj*), خوشنما (*khushnuma*), ناخوش (*nakhush*) can be added with خوش (*khush*) creating different variant forms of the word. In the following subsection, a brief explanation is provided on the usage of noun, verb, adjective, adverbs in Urdu language.

*A. Noun*

In Urdu, nouns are inflected in number and case and have an inherent gender. There are some types defined for case, number and gender.

- Case: Nominative/Oblique/Vocative.
- Number: Singular/Plural.
- Gender: Masculine/Feminine.

Nouns can be divided into 15 groups on the basis of inflection. From those 15 groups, 1 group is based on the Masculine and other are based on the Feminine.

For Example:
Singular masculine nouns ending with (ا {alif}, a), (ہ {he}, h) and (ع {ain}, e):
This group also includes the Arabic loan nouns ending with (ہ {he}, h). According to this rule, first, if a word ends with letter (ا {alif}, a) or (ہ {he}, h) then:

- To make plural nominative and singular oblique, the last letter is replaced by letter (ے {ye}, E).
- To make plural oblique, the last letter is replaced by string (وں {on}, wN).
- To make plural vocative, the last letter is replaced by letter (و (*wow*), w).

Second, if a word ends with (ع {*ain*}, e) then the rules will remain same as above except that the above mentioned letters will be added at the end of words without replacing any existing letter. Table 1 displays examples from this group.

TABLE I
An Example of Noun Group

|  | Nominative | Oblique | Vocative |
|---|---|---|---|
| Singular | بتھوڑا (*hathoda*) | بتھوڑے (*hathode*) | بتھوڑے (*hathode*) |
| Plural | بتھوڑے (*hathode*) | بتھوڑوں (*hathodon*) | بتھوڑو (*hathodo*) |

There are some more inflections present with masculine and feminine cases:

- Singular masculine nouns ending with (ان {*aan*}, an). E.g. دھواں (*dhuan*).
- Singular feminine nouns ending with (ی {*i*}, y). E.g. لڑکی (*ladki*).
- Singular feminine nouns ending with (ا {*alif*}, a), (ان {*aan*}, an), (وں {*on*}, on). E.g. کرسیاں (*kursiyan*).

*B. Verb*

The Urdu verb are very much complex as compared to other word classes. In terms of Morphology, Urdu verb inflects in different form:

- Gender: Masculine, Feminine.
- Number: Singular, Plural.
- Person: First, Second (casual, familiar, respectful), Third.
- Mood & Tense: Subjunctive, Perfective, Imperfective.

Urdu verb presents direct and indirect causative behavior. Generally, for every verb, one stem or root word is formed. It could be Intransitive, transitive etc. when this basic stem forms then two other forms (direct & indirect causatives) makes for that verb. These three forms are actually regular verbs. For example, consider a verb: کر (*kar*-To Do)

- Infinitive form: کرنا (*karna*)
- Direct causative infinitive form: کرانا (*karana*)
- Indirect causative infinitive form: کروانا (*karvana*)

کرنا (*karna*), کرانا (*karana*) and کروانا (*karvana*) are three regular verbs and inflect in tense, mood, aspect, gender and number.

*C. Adjective*

Urdu adjectives are also inflected on the basis Gender, number and case marking. E.g. the singular direct 'فرتیلا' (*furteela*-Active) becomes 'فرتیلے' (*furteele*) in all other masculine cases and 'فرتیلی' (*furteeli*) in all feminine case. Therefore no need to add new affixes for adjectives.

*D. Adverb*

Adverbs may be in variable forms or invariable forms like an adjectives. It depends upon whether they change the form with the noun or with the verb. We can divide them into following categories as mentioned below:

- Adverbs of Time: روجانہ (*rojanah*-Daily), اکثر (*aksar*-Usually)
- Adverbs of Place: یہاں (*yaha*-Here), وہاں (*vaha*-There)
- Adverbs of Manner: یکایک (*yakayak*-Suddenly)
- Adverbs of Degree: چھوٹا (*chhota*-Small), لمبا (*lamba*-Long).

IV. PROPOSED SYSTEM

In our proposed system, we have created a Rule Based Stemmer for Urdu. Here we have created an affix list of 119 rules. It contains 107 postfix or suffix and 12 prefix. These affixes are used for extracting the root word or stem.

The below list does not contain the complete affix list. Some more affixes can also be added, but if we are adding some more affixes then we might face the problem of Over Stemming, for this reason we have not add any other affix. In this rule based approach, we have ordered the affix list in descending order in respect to length. This system is based on Stripping Method where the affixes are just removed. The affixes with long length are removed first and further if required then the affix with shorter length is removed. In some of the cases we don't find the correct stem. For Example word حیات (*hayaat*) should not have been reduced but it reduces in ہے (*hay*) root form and also separates the suffix ات (*aat*). This root ہے (*hay*) does not have a meaning. Table 2 shows some of the resultant stem or root word generated from our proposed Stemming algorithm.

| انیں | یں | ی | و | ے |
|---|---|---|---|---|
| تے | نے | ین | ات | ا |
| ائی | وں | یاں | تا | یوں |
| ں | بد | نو | ناک | گے |

Fig. 1 Suggestive Affix List

TABLE II
RESULT OF THE STEMMING ALGORITHM

| Word | Prefix | Stem | Suffix |
|---|---|---|---|
| علاقوں (*ilaaq-aun*) | | علاقہ (*ilaaqah*) | وں (*aun*) |
| فاصلے (*faasl-ee*) | | فاصلہ (*faasalah*) | ے (*ae*) |
| سوالات (*saval-aat*) | | سوال (*savaal*) | ات (*aat*) |
| لڑکیاں (*ladki-yaan*) | | لڑکی (*ladki*) | یاں (*yaan*) |
| راجویر (*raaj-veer*) | راج (*raaj*) | ویر (*veer*) | |
| نوجوان (*nau-javaan*) | نو (*nau*) | جوان (*javaan*) | |
| لاجواب (*laa-javaab*) | لا (*la*) | جواب (*javaab*) | |
| بدنصیب (*bad-naseeb*) | بد (*bad*) | نصیب (*naseeb*) | |

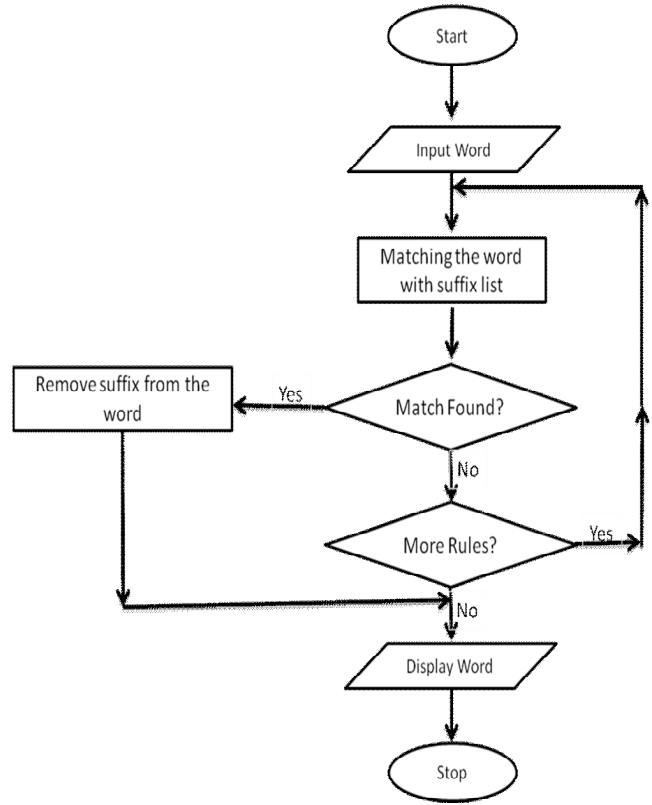

Fig. 2 Rule Based Stemmer for Urdu

Figure 2 shows the working of our Stemming algorithm. In this algorithm, we collected the affixes and found the correct matching of affixes with their corresponding word. If we get a match then the respective affix is detached from the respective word. If the match is not found then the same word is displayed as it is.

Our approach provides us correct output, so we can say that our system can show results which higher accuracy. The Results and Evaluation of our System is presented in next Section.

## V. EVALUATION

With this system we wanted to know that how much accurate output does it give us. It means that we were concerned about the accuracy of our system. For calculating the accuracy we have used the following formula:

$$Accuracy\ (\%) = \frac{Accurate\ stemmed}{Total\ no\ of\ given\ word} \times 100 \quad (1)$$

In this paper, we have tested the system by executing the proposed algorithm on the test data of 2000 words. Table 3 shows Summary for test data.

TABLE III
SUMMARY FOR TEST DATA

| Test Data Features | Total Count |
|---|---|
| Total Words | 2000 |
| Correct stemmed output | 1730 |
| Wrong output | 270 |
| Unique Words | 167 |
| Min Length | 4 |
| Max Length | 15 |

Here we have checked our system on the test data of 2000 words. Among these 2000 words, 1730 gives correct stem and 270 gives incorrect stem. From these 1730 words, 167 are unique words. This shows that words are not matched with the affix list and in return displays the same word. Due to the problem of over stemming and under stemming, we received 270 wrong stems.

By using accuracy formula (1) and on the basis of given test data we achieved an accuracy of 86.5%. Figure 3 shows the result of test data.

## VI. CONCLUSION

In this paper, we have shown the design and implementation of Rule Based Stemmer for Urdu. It is useful for complex and morphologically variant words. This rule based stemmer is capable of capturing the root word and separates the affixes. We have evaluated our system and obtained 86.5% accuracy.

Further the work will be done to increase the accuracy of Urdu Stemmer, in which, the affix list will be improved. Later, we will use ripple down approach to generate some rules for exceptional words which occur due to the problem of under stemming and over stemming.

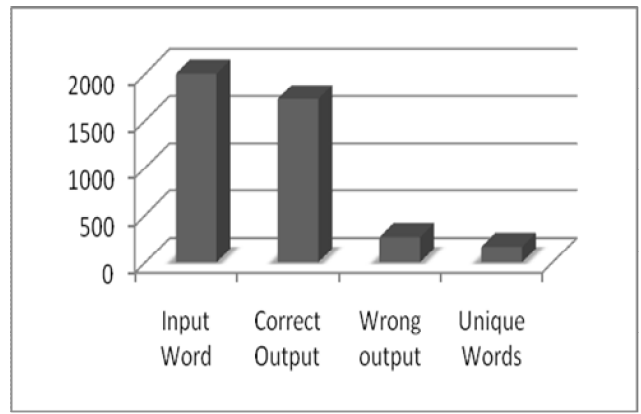

Fig. 3  Result of Test Data